\pdfoutput=1

\documentclass[11pt]{article}

\usepackage{acl}

\usepackage{times}
\usepackage{latexsym}

\usepackage[T1]{fontenc}

\usepackage[utf8]{inputenc}

\usepackage{xspace}

\usepackage{hyperref}
\usepackage{url}
\usepackage{comment}
\usepackage{amsmath}
\usepackage{amssymb}
\usepackage{algorithm}
\usepackage{algorithmic}
\usepackage{varwidth}
\usepackage{colortbl}
\usepackage{color}
\definecolor{color2}{RGB}{99, 145, 122}
\definecolor{color3}{gray}{0.9}
\usepackage{makecell}
\usepackage[para]{threeparttable}
\usepackage{booktabs}
\usepackage{graphicx}
\usepackage{multirow}

\newcommand{\dataset}{{OCKL}\xspace}
\usepackage{subfigure}
\usepackage{marginnote} 
\usepackage{tcolorbox}
\usepackage{xcolor}

\definecolor{lightgray}{gray}{0.7}
\title{Online Continual Knowledge Learning for Language Models}

\author{Yuhao Wu$^1$\footnotemark[1], Tongjun Shi$^1$\footnotemark[1], Karthick Sharma$^2$\footnotemark[2], Chun Wei Seah$^1$, Shuhao Zhang$^3$\footnotemark[3] \\
  $^1$Singapore University of Technology and Design; $^2$University of Sri Jayewardenepura; \\ $^3$Nanyang Technological University\\ 
  \texttt{\{yuhao\_wu, chunwei\_seah, tongjun\_shi\}@sutd.edu.sg} \\ \texttt{en93899@sjp.ac.lk} \\ \texttt{shuhao.zhang@ntu.edu.sg}
}

\begin{document}
\maketitle
\renewcommand{\thefootnote}{\fnsymbol{footnote}} 
\footnotetext[1]{The first two authors contributed equally.} 
\footnotetext[2]{Work done while the author was interning at SUTD.}
\footnotetext[3]{Corresponding author.}

\begin{abstract}
Large Language Models (LLMs) serve as repositories of extensive world knowledge, enabling them to perform tasks such as question-answering and fact-checking. However, this knowledge can become obsolete as global contexts change.
In this paper, we introduce a novel problem in the realm of continual learning: \textit{Online Continual Knowledge Learning} (\textbf{OCKL}). This problem formulation aims to manage the dynamic nature of world knowledge in LMs under real-time constraints. We propose a new benchmark and evaluation metric designed to measure both the rate of new knowledge acquisition and the retention of previously learned knowledge. Our empirical evaluation, conducted using a variety of state-of-the-art methods, establishes robust baselines for OCKL. Our results reveal that existing continual learning approaches are unfortunately insufficient for tackling the unique challenges posed by OCKL. We identify key factors that influence the trade-off between knowledge acquisition and retention, thereby advancing our understanding of how to train LMs in a continually evolving environment.

\end{abstract}

\section{Introduction}
Large Language Models like LLaMa2~\citep{touvron2023llama} and GPT-3~\citep{brown2020language} have demonstrated remarkable proficiency in Knowledge Intensive Language Tasks (KILT)~\citep{petroni2021kilt}. Despite their success, a significant limitation is their struggle to adapt to rapidly evolving global knowledge, as highlighted by their performance in the LAnguage Model Analysis (LAMA) framework~\citep{petroni2019language}. The conventional method of augmenting LLMs with an external, continually updated knowledge base introduces latency and reduces the models' capacity for real-time contextual understanding~\citep{raffel2019exploring}.


\begin{figure}[t]
\includegraphics[width=0.45\textwidth]{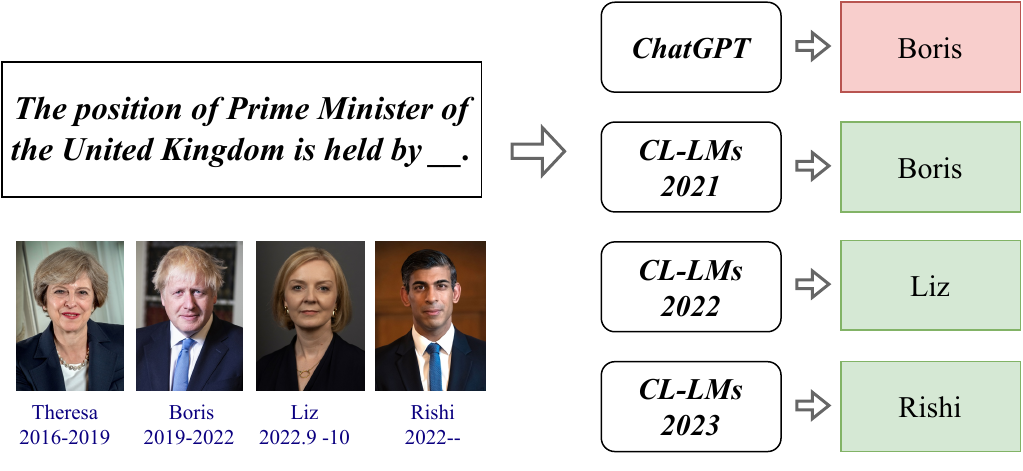}
    \caption{\small Static models like T5~\citep{2020t5}, due to their knowledge cutoff, cannot provide timely answers to time-variant knowledge. 
    }
    \label{fig:CL}
\end{figure}

As illustrated in Figure~\ref{fig:CL}, current LLMs often yield outdated or incorrect answers when confronted with time-sensitive questions. 
Addressing these constraints necessitates a framework that enables real-time, internal knowledge updating in LLMs. While some studies have begun addressing this need~\citep{mitchell2021fast, de2021editing, zhu2020modifying, dhingra2022time, jang2021towards}, they have not fully resolved the challenges associated with rapid LLM updating.

To tackle these issues, we propose the \textit{Online Continual Knowledge Learning} (OCKL) framework. OCKL differentiates itself from existing Continual Knowledge Learning (CKL) by prioritizing immediate, ongoing internal knowledge updates within significantly shorter timeframes, ranging from days to mere seconds. This approach, divergent from offline methods, requires single-pass updates (epoch$=$1) due to the high velocity and volume of incoming data~\citep{aljundi2019gradient}. OCKL presents a novel challenge in LM updates, for which a robust benchmark is essential yet currently absent.

To evaluate LMs within our framework, we introduce two new metrics: \textsc{Knowledge Acquisition Rate (KAR)} and \textsc{Knowledge Gap (KG)}, assessing text processing efficiency, knowledge acquisition, retention, and real-time learning variances. Additionally, we examine the suitability of existing continual learning methods for OCKL, exploring model architectures and training methodologies that mitigate knowledge ``forgetting''~\citep{he2021analyzing, hu2021lora}. We also utilize advanced coreset selection techniques to enhance knowledge acquisition and reduce computational load~\citep{yoo2019learning, li2022camel}. This comprehensive approach establishes a robust baseline for the OCKL framework and lays the foundation for future research aimed at optimizing the adaptability and efficiency of LMs in real-time data environments.


\textbf{Contributions.} This paper presents multifaceted contributions:
\begin{enumerate}
    \item \textbf{Novel Framework and Benchmark}: The introduction of \textbf{OCKL}, a new framework designed to better understand the shortcomings of current LM updating practices facing high velocity data streams. 
    \item \textbf{Baseline Approaches and Metrics}: Innovative metrics like Knowledge Acquisition Rate and Knowledge Gap are introduced for nuanced evaluation of LMs' learning conditions.
    \item \textbf{Empirical Insights and Challenges}: Exhaustive analyses within the OCKL benchmark reveal crucial insights and challenges. For instance, rehearsal methods consistently outperform other approaches despite buffer inefficiencies.
\end{enumerate}

\section{Related Work}
Language Models (LMs), particularly when augmented with external knowledge sources like Retrieval-Augmented Generation (RAG)~\citep{lewis2020retrieval,guu2020retrieval,ram2023context}, aim to stay current by updating these sources or querying the internet for the latest information. However, these methods grapple with challenges, notably ``hallucination'', where models generate incorrect information as factual, a problem amplified in larger LMs~\citep{zhang2021situatedqa,longpre-etal-2021-entity}.


While retraining language models from scratch with updated data offers a comprehensive solution, it is computationally intensive and environmentally unsustainable~\citep{patterson2021carbon}. A more feasible alternative, incremental pretraining on smaller, updated corpora, faces the risk of ``catastrophic forgetting'', where newly acquired knowledge supersedes prior learning~\citep{de2021continual}. Existing mitigation strategies such as regularization~\citep{buzzega2020dark, caccia2021new}, rehearsal-based methods~\citep{chaudhry2019continual,aljundi2019gradient, yoon2021online, chrysakis2020online,chaudhry2021using}, and parameter expansion~\citep{rusu2016progressive, hu2018overcoming, mallya2018packnet} are primarily designed for offline contexts, often requiring multiple passes through data and thus not aligning with real-time, online scenarios. 

In the specific realm of CKL for LMs, various approaches like those proposed by \citep{jang2021towards,luu2021time,dhingra2022time,margatina2023dynamic} have been developed to update and retain knowledge over time. However, similar to the broader CL methodologies, these CKL strategies are predominantly offline, iterating over the same data multiple times and lacking in their ability to swiftly adapt to the dynamic and evolving nature of streaming data contexts. 
The limitations of these traditional methods in the context of our framework are further discussed in Appendix~\ref{tranditional CL}.

\section{Online Continual Knowledge Learning} 
In this section, we will provide the illustrations of the task formulation, conduct dataset analysis, and present the proposed metrics for online continual knowledge learning.


\begin{figure}[!t]
\centering
    \includegraphics[width=0.4\textwidth]{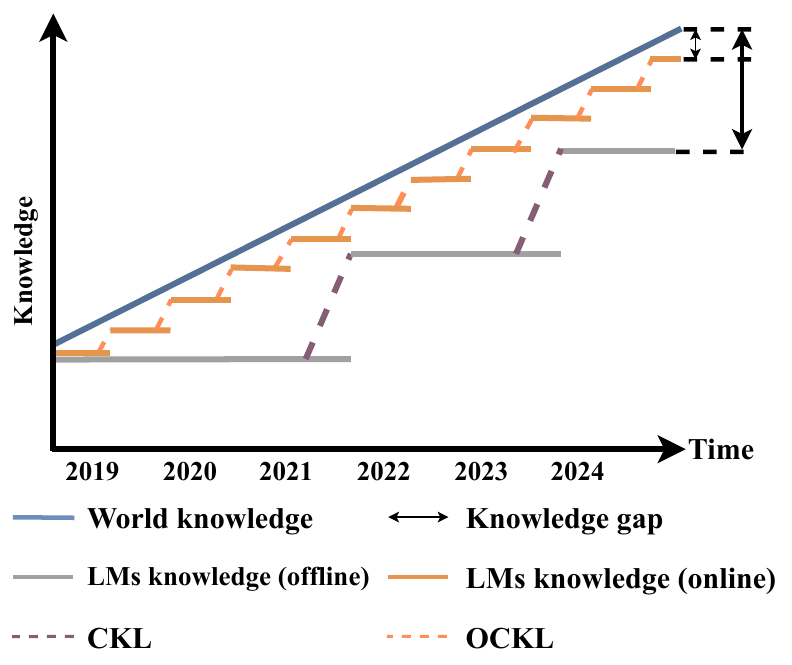}
    \caption{Comparative Analysis of Knowledge Synchronization in Offline (CKL) and Online (OCKL) Learning Paradigms. 
}
    \label{fig:overview}
\end{figure}
\subsection{OCKL Formulation}

\paragraph{Motivation and Challenges}
The evolution of real-world information necessitates the continual update of LMs. Despite the capacity of LMs to act as reservoirs of global knowledge through pretraining on extensive corpora, this knowledge is susceptible to obsolescence. This leads to the formulation of a novel Continual Learning problem--\textit{Online Continual Knowledge Learning}, which aims to harmonize the acquisition of new knowledge with the preservation of existing information under real-time constraints.
Figure~\ref{fig:overview} contrasts the shortcomings of traditional Continual Knowledge Learning methods~\citep{jang2021towards} with the promise of OCKL. CKL allows LMs to accumulate extensive knowledge but does not offer a mechanism for real-time adaptation. OCKL seeks to rectify this by facilitating frequent, incremental updates to the LM, enabling it to adapt to dynamically changing information landscapes.

\paragraph{Objectives and Scope}
Against the challenges intrinsic to OCKL, we outline two key objectives that guide this study: 1) to establish a real-time evaluation framework with a comprehensive suite of metrics to measure the rate of new knowledge acquisition and the effectiveness of existing knowledge retention in LMs; 2) to refine LM training protocols with a focus on real-time knowledge updates. We employ coreset selection techniques to optimize both speed and efficiency in knowledge acquisition.

\paragraph{Formal Framework}
The continual updating of LMs is represented as transitions between model states: \( \theta_{t-1} \rightarrow \theta_t \). Knowledge updates are sourced from a data stream \( \left\{ X_t^i \right\}_{i=1}^{n} \sim D_{t} \), and factual verification is performed via a QA stream \( \left\{(Q_t^i, A_t^i) \right\}_{i=1}^{n} \sim D_{t} \). At each time step \( t \), two activities are conducted: 
\begin{enumerate}
    \item Updating: the model \( \theta_t \) is updated using \( \left\{ X_t^i \right\}_{i=1}^{n} \sim D_{t} \). 
    \item Evaluating: the model \( \theta_t \) is evaluated through the QA stream, generating answers \( \left\{ \widetilde{A_t^i} \right\}_{i=1}^{n} \) for questions \( \left\{ Q_t^i \right\}_{i=1}^{n} \).
\end{enumerate}
\subsection{Efficiency-Driven Coreset Selection in OCKL}
\label{subsec:coreset}
Coreset selection is a strategy aiming at isolating a representative and informative subset from a comprehensive training dataset. Its utility lies in enabling models to approximate the performance achieved when trained on the complete dataset. In the context of rehearsal-based continual learning, coreset selection has previously been employed as a mechanism to counteract catastrophic forgetting through the retention of a knowledge buffer~\citep{tiwari2022gcr, yoon2021online}. 

In this work, we enhance the application of coreset selection, specifically targeting the acceleration of OCKL. We define a subset \( S \subseteq \{1, 2, \ldots, n\} \), which consists of indices that select specific elements \( X_t^i \) from the original dataset. Consequently, the chosen subset from \( \left\{ X_t^i \right\}_{i=1}^{n} \) is denoted as \( \left\{ X_t^i : i \in S \right\} \).
By selecting pertinent data samples, our approach seeks not only to preserve antecedent knowledge but also to alleviate the computational demands inherent in the training process, thereby contributing to more efficient online learning systems. The ensuing discussion will explore three specific methods for coreset selection—Random Sampling, K-Center selection, and Model-Based~\citep{yoo2019learning} selection—each with its advantages and limitations in the context of OCKL. These methods are illustrated further in Appendix~\ref{coreset}.

\subsection{Dataset Characteristics and Methodology}

\paragraph{Overview}
To facilitate the construction of an adaptive Language Model capable of processing time-sensitive knowledge, our dataset is divided into two primary components: the \textbf{Knowledge Stream} and the \textbf{QA Stream}. These streams are generated based on Wikidata's exhaustive knowledge base, spanning from 2019 to 2023. The dataset features both time-invariant and time-variant facts, parsed and structured through SLING\footnote{A natural language frame semantics parser by \href{https://ringgaard.com/knolbase}{Ringgaard Research}}.

\paragraph{Data Preprocessing \& Construction}
The primary data source, a raw Wikidata dump, is initially mapped onto data frames. These are subsequently refined to distill facts related to specific entities. Facts are then categorized based on their property attributes, separating them into time-invariant and time-variant types. A detailed visual representation of this preprocessing methodology is available in Appendix~\ref{appen:datasets}. We construct \textbf{Knowledge Stream} by integrating the refined facts with corresponding property templates and generating a \textbf{corpus} along with an associated date for each entry. Likewise, the \textbf{QA Stream} is constructed in alignment with established research methodologies~\citep{margatina2023dynamic, dhingra2022time}, comprising the \textbf{query}, \textbf{answer}, and \textbf{date}.


\paragraph{Data Statistics Analysis}
Following the construction of the data streams, we present key statistical attributes. The Cumulative Distribution Function (CDF) of the token and date changes over time is shown in Appendix~\ref{append:analysis}.
The Knowledge Stream contains 94,568 data points, with average text and token lengths of 70.1 and 12.7, respectively. It comprises 37.6\% time-invariant facts and 62.4\% time-variant facts. For the latter, the average text and token lengths stand at 74.8 and 13.7, respectively.
For a comparative evaluation, we further refine our initially constructed datasets to eliminate redundancies, ensuring that each fact is represented only once. This non-redundant version includes 1,929,045 texts, with average token and text lengths of 11.24 and 62.54, respectively. Notably, 99.97\% of these texts are time-variant facts, while a mere 0.03\% are time-invariant.

\subsection{Evaluation Metrics}
For a comprehensive and unbiased assessment of the OCKL problem, we have devised two novel performance metrics and refined the standard BWT (Backward Transfer) and FWT (Forward Transfe) approaches, together with EM (Exact Match). The details of EM, BWT and FWT are provided in Appendix~\ref{other Metrics}.

\paragraph{Knowledge Acquisition Rate (KAR)\label{LearningSpeed}} 
The assessment of a model's knowledge acquisition rate is nontrivial due to its dependency on various factors. Multiple factors contribute to the varying speeds at which different methods process tokens. Key contributors include the volume of data processed (rehearsal-based), differences in the number of trainable parameters (expansion parameter-based), and additional computations required (regularization-based). These intricacies are further compounded by divergences in knowledge acquisition and retention capabilities across different learning algorithms. To navigate these complexities, we introduce a comprehensive formula for quantifying the effective learning rate of a model. We leverage previously defined metrics: Forward Transfer quantifies the incorporation of new knowledge, while Backward Transfer serves as an index of knowledge retention or forgetfulness. The sum of these metrics provides a net knowledge gain per learning iteration. Accordingly, the genuine knowledge acquisition rate can be calculated using the following equation:
\begin{equation}
\small
\text{KAR} = ( \text{{FWT}} + \text{{BWT}} ) \times \frac{{\text{{Total Tokens}}}}{{\text{{Training Time}}}}
\end{equation}

\paragraph{Knowledge Gap (KG)\label{KG}}
The Knowledge Gap metric provides a tangible measure of knowledge dynamics within a Language Model by employing vector representations and distance metrics. We leverage the concept of parametric knowledge, which asserts that an LM encodes world knowledge within its parameters~\citep{izacard2022few, jang2021towards} for our evaluation.

\textit{Operationalization:} 
As a case study, consider the T5 model. Its embedding layer vectorizes input data before the encoding phase, offering a representation of ``world knowledge'' denoted as \(T_w\). Further, the final encoder layer extracts hidden states from the input query, encapsulating ``model knowledge'' denoted as \(T_m\).

\textit{Metric Formulation:}
We compute the KG metric as a similarity distance between two sequences, \(T_1\) and \(T_2\), which can represent either world or model knowledge. The framework allows for the following configurations:

\begin{enumerate}
  \item When \(T_1 = T_w\) and \(T_2 = T_m\), the Euclidean distance \(d(T_{1}, T_{2})\) quantifies the alignment between model knowledge and world knowledge. A smaller value indicates better fidelity.
  
  \item When \(T_1 = T_{m_{(t-1)}}\) and \(T_2 = T_{m_{(t)}}\), where \(t\) denotes the time stamp, \(d(T_{1}, T_{2})\) evaluates the \textbf{\textit{forgetting}} behavior, capturing knowledge attrition during online learning.
  
  \item When \(T_1 = T_{m_{(t+1)}}\) and \(T_2 = T_{m_{(t)}}\), \(d(T_{1}, T_{2})\) measures the \textbf{\textit{updating}} behavior, assessing knowledge acquisition or refinement during online learning.
\end{enumerate}

By applying the KG metric in these configurations, we offer an assessment of the model's capacity for knowledge retention and acquisition, which is particularly vital for evaluating models deployed in dynamic, real-world scenarios.

\subsection{Continual Learning Method}
\label{CL method}
In this paper, we have selected a variety of CL methods for OCKL, specifically including:

\textbf{Vanilla} refers to a standard setup utilized for additional training, as discussed by~\citet{gururangan2020don}. 

\textbf{RecAdam}~\citep{chen2020recall} falls into the category of regularization methods. It differs from traditional regularization methods like EWC~\citep{kirkpatrick2017overcoming} by imposing a more stringent independence assumption among model parameters.

\textbf{Knowledge Distillation}~\citep{gou2021knowledge} is to speed up model inference by minimizing the representation gap between two models. 

\textbf{Mix-Review}~\citep{he2021analyzing} falls into the category of rehearsal methods, which assumes access to the initial pretraining corpus and mixes in random subsets of the initial pretraining data during continued pertaining. 

\textbf{LoRA}~\citep{hu2021lora} is a parameter-expansion technique presented by~\citet{hu2021lora}. Instead of updating the original LM parameters, it introduces trainable rank-decomposition matrices in each layer for continued pertaining.

\textbf{K-Adapter}~\citep{wang2020k} is another parameter expansion method that freezes the original LM parameters while adding a \textit{k} number of new layers, namely \textit{adapters}, that are updated during continued pre-training. 

\textbf{Modular}~\citep{jang2021towards} is a novel parameter expansion strategy tailored for encoder-decoder models. The original pretrained encoder is preserved, but a supplementary, randomly initialized encoder is introduced for updates during the next phase of training.

\section{Experiments}
In this section, we will begin by presenting our experimental setup followed by an in-depth discussion of the outcomes from our experiments. 


\subsection{Experimental Setup}
We conducted extensive experiments using the T5 encoder-decoder model~\citep{raffel2019exploring}. This Language Model, with around 737M parameters, was originally pretrained on the C4 dataset from April 2019 and the Wikipedia dataset from May 2020. For our analysis of the OCKL benchmark, we established several baseline methods. These methods are categorized into three main groups: \textit{regularization} (i.e., RecAdam, KD as Knowledge-Distillation), \textit{rehearsal} (i.e., Mix-Review), and \textit{parameter-expansion} (i.e., LoRA, Kadapter, and Modular). The detailed descriptions of the CL setups and our hyper-parameters used for each method are provided in Appendix~\ref{CL setups} and \ref{Hyper-parameter} respectively.

\begin{table*}[t]
\centering
\begin{minipage}[t]{0.59\textwidth}
    \centering
    \large
    \addtolength{\tabcolsep}{-0pt}
    \fontsize{9}{12}\selectfont   
    \raisebox{\dimexpr-\height+\ht\strutbox\relax}{
    \resizebox{\textwidth}{!}{
    \begin{tabular}{clcccccc}
    \toprule
    \multicolumn{1}{c}{\textbf{Corpus}} & \multicolumn{1}{c}{\textbf{Method}} & \textbf{\makecell{\# of Params \\ (Trainable / Total)}} & \makecell{EM (\%) $\uparrow$} &
    \makecell{KG $\downarrow$} & \makecell{KAR $\uparrow$} \\   \midrule
    \multirow{8}{*}{\textsc{Redundancy}} & T5-Initial            & 0M / 222M     & \textcolor{white}{0}0.50  & 91.601 & --         \\ \cmidrule{2-6}
    &T5-Vannila            & 222M / 222M   & \underline{21.41} & 91.792 & \underline{3997.07}  \\
    &T5-RecAdam            & 222M / 222M   & 20.48 & 91.634 & 3282.73 \\
    &T5-Mix-Review         & 222M / 222M   & \textbf{23.26} & 91.802 & \textbf{4264.82}  \\
    &T5-LoRA               & 113M / 223M   & 16.42 & \underline{91.453} & 2031.78 \\
    &T5-K-Adapter   & 141M / 251M   & 15.21 & 91.601 & 1711.34 \\
    &T5-Modular            & 198M / 307M   & 13.61 & 91.601 & \textcolor{white}{0}980.80 \\
    &T5-KD                & 222M / 222M   & 15.02  &\textcolor{white}{0}\textbf{89.38}  & \textcolor{white}{0}669.04\\
    \midrule
    \multirow{8}{*}{\textsc{Redundancy-free}} & T5-Initial            & 0M / 222M     & 0.34  & 16.571 & -- \\ \cmidrule{2-6}
    & T5-Vannila            & 222M / 222M   & \underline{15.09} & 16.769 & \textbf{4471.01}\\
    & T5-RecAdam            & 222M / 222M   & 14.03 & 16.729 & 3240.77\\
    &T5-Mix-Review         & 222M / 222M   & \textbf{15.38} & 16.697 & \underline{4105.50}\\
    &T5-LoRA               & 113M / 223M   & 14.53 & 16.768 & 3363.83\\
    &T5-K-Adapter   & 141M / 251M   & 12.43 & \underline{16.571} & 3699.39\\
    &T5-Modular            & 198M / 307M   & 13.39 & \underline{16.571} & 3343.83\\
    &T5-KD                 & 222M / 222M   & \textcolor{white}{0}7.09  & \textbf{12.480}  & \textcolor{white}{0}267.30\\
    \bottomrule
\end{tabular}
    }}
    \caption{\small Redundant and redundant-free datasets results. Different continual learning strategies come with their own set of limitations due to the constraints set by online learning and the dense emergence of new information in data streams. $\downarrow$ represents the lower the better; $\uparrow$ represents the higher the better. The best results for each task and metric are shown in bold. The second-best results are underlined with a horizontal line.}
    \label{table:main}    
\end{minipage}%
\hfill
\begin{minipage}[t]{0.39\textwidth}
    \centering
    \large
    
    \addtolength{\tabcolsep}{-0pt}
    \fontsize{9}{12}\selectfont
    \raisebox{\dimexpr-\height+\ht\strutbox\relax}{
    \resizebox{\textwidth}{!}{
    \begin{tabular}{lcccc}
\toprule
\multicolumn{1}{c}{Method} & EM $\uparrow$ & BWT $\uparrow$ & FWT $\uparrow$ & KG $\downarrow$ \\ \midrule 
T5-Vanilla & 15.09 & -4.18 & 9.27 & 16.769 \\
T5-RecAdam & 14.03 & -4.48 & 8.98 & 16.729 \\
T5-Mix-Review & \textbf{15.38} & -4.49 & \textbf{9.28} & 16.697 \\
T5-LoRA & 14.53 & -4.34 & 8.46 & 16.768 \\
T5-Kadapter (k=2) & 12.43 & -3.59& 7.58 & 16.576 \\
T5-Modular & 13.39 & -3.96 & 7.97 & 16.570 \\
T5-KD & \textcolor{white}{0}7.09 & \textbf{-2.82} & 4.96 & \textbf{12.480} \\ \bottomrule
\end{tabular}%
    }}
    \caption{\small Comparison of knowledge forgetting and updating for different methods.}
    \label{tab:analysis}   
    \vspace{\baselineskip}
    \centering
    \large
    \fontsize{9}{12}\selectfont
    \raisebox{\dimexpr-\height+\ht\strutbox\relax}{
    \resizebox{\textwidth}{!}{
    \begin{tabular}{l@{\extracolsep{\fill}}ccccccc}
    \toprule
    \multicolumn{1}{c}{\textbf{Method}} & \textbf{\makecell{\# of Params \\ (Trainable / Total)}} & \makecell{EM (\%) $\uparrow$} &
    \makecell{KG $\downarrow$} & \makecell{KAR $\uparrow$} \\ \hline
    T5-initial & 0M / 737M & \textcolor{white}{0}0.49 &  7.11 & - \\
    \midrule
    T5-Vanilla & 737M / 737M& 16.19 & 7.02 & 2378.68  \\
    T5-RecAdam & 737M / 737M  & 14.85 & 7.04 & 1840.27\\
    T5-Mix-Review & 737M / 737M  & 16.18 & 7.01 & 2658.96 \\
    T5-LoRA   & 403M / 738M & \textbf{17.17} & 7.12 & \textbf{2761.98} \\
    T5-Kadapter (k=2) & 427M / 762M  & \textcolor{white}{0}6.71  & 7.11 & 1542.39 \\
    T5-Modular   & 438M / 773M & \textcolor{white}{0}8.55  & 7.11 & 1765.85 \\
    T5-KD    & 737M / 737M & \textcolor{white}{0}6.30  & \textbf{5.50} & \textcolor{white}{0}122.95 \\
    \bottomrule
\end{tabular} 
    }}
    \caption{\small The experiment results of T5-large model in the redundancy-free stream data.} 
\label{table:t5_large}
\end{minipage}
\end{table*}

\subsection{Experiments Result}
\subsubsection{Main Result}

Table~\ref{table:main} presents the outcomes of our experiments carried out on the \dataset benchmark. We concentrated on three pivotal metrics: Exact Match (EM), Knowledge Gap (KG), and Knowledge Acquisition Rate (KAR). The T5 models underwent an initial pretraining phase using the C4 dataset, which entailed roughly 1 trillion token updates, along with Wikipedia data up until 2019. This served as our foundational model for a series of continuous learning tests.

We initiate our analysis using a dataset containing a considerable volume of redundant data, reflecting the redundancy commonly found in real-world textual streams. As delineated in Table~\ref{table:main} under the \textit{Redundancy} subsection, the method named T5-Mix-Review excels in EM and KAR compared to alternative approaches. In contrast, the parameter expansion method e.g. T5-LoRA, performs admirably in CKL~\citep{jang2021towards} but falters in OCKL. This performance drop could be ascribed to T5-LoRA's restricted trainable parameters, as it freezes the primary network parameters, limiting both retention and learning capabilities. In the \textit{Redundancy-Free} section of Table~\ref{table:main}, both the naive approach T5-Vanilla and the rehearsal-based method T5-Mix-Review exhibit efficacy in learning from the online knowledge stream, as evidenced by their high EM scores. This observation underscores the varying effectiveness of rehearsal methods in CL, parameter-expansion methods in CKL, and their overall limitations in our OCKL setup. The only exception is the rehearsal method, which surpasses the simple T5-vanilla approach. Although T5-KD demonstrates commendable performance, as highlighted by \citet{jin2021lifelong}, it lags in OCKL due to the time-intensive nature of distillation-based methods. To achieve superior performance, one must rely on the teacher model's outcomes. Furthermore, as T5-KD loss influences the embeddings of both the teacher and student models directly, it registers the lowest score in the Knowledge Gap metric.

We extend our evaluation by scrutinizing various CL methods in Table~\ref{tab:analysis}, focusing on redundancy-free stream data. Remarkably, as indicated in this table, both T5-Kadapter and T5-Modular methods—attributable to their frozen encoder components—register an identical KG metric when compared to their original model. It is evident from Table~\ref{tab:analysis} that no single method excels across all metrics. Our analysis suggests that regularization methods outperform others in mitigating the phenomenon of knowledge forgetting in an online learning context. Conversely, parameter expansion methods uniformly show suboptimal performance, as reflected in their notably weaker Forward Transfer scores, possibly due to the constraints imposed by their set of trainable parameters. Rehearsal-based methods exhibit robust FWT results which is the determinant in their superior Exact Match performance. Of all the evaluated methods, the T5-KD approach yields the smallest Backward Transfer, aside from its relatively lower EM, which may imply its aptness for offline continual learning scenarios where Epoch $>$ 1.
\subsubsection{T5-large Experiment Result}
\label{appen:t5_large}
Table~\ref{table:t5_large} presents the experimental results for larger model scales in T5-large, and the outcomes were unexpectedly impressive to a certain extent. Most methods, including T5-Vanilla and T5-Mix-Review, produced results consistent with those shown in Table \ref{table:main}, maintaining high performance according to the FWT and KAR metrics. Interestingly, the performance of T5-loRA~\citep{hu2021lora} significantly outperformed that of the T5-base experiments. Surprisingly, other methods for parameter expansion, such as T5-Kadapte and T5-Modular, did not yield corresponding improvements. This suggests that the enhancement achieved by loRA cannot be solely attributed to an increase in the number of trainable parameters. One potential explanation for this phenomenon may stem from LoRA's unique characteristics when contrasted with other parameter expansion techniques. LoRA's distinctiveness is manifested through the incorporation of trainable rank-decomposition matrices in every layer throughout the continuous pretraining process.


\subsubsection{Impact of Coreset Selection Strategies on OCKL}
\begin{table*}[t]
\centering
\small
\fontsize{8.5}{12}\selectfont
\label{tab:cs}
\resizebox{0.8\textwidth}{!}{%
\begin{threeparttable}
\begin{tabular}{clccccccccc}
\toprule
\multicolumn{1}{c}{\textbf{Corpus}} & \multicolumn{1}{c}{\textbf{Method}} & \textbf{EM} & \textbf{BWT} & \textbf{FWT} & \textbf{KG} & \textbf{Forget} & \textbf{Update} & \textbf{KAR} \\ \midrule
\multirow{4}{*}{\textsc{Redundancy}} 
& w/o selection          & 21.41 & 0.59  & 0.64 & 91.79 & 0.17 & 0.17 & 3997.07\\ 
\cmidrule{2-9}
& Random                 & \textcolor{white}{0}9.44  & 0.16  & 0.16 & 91.48 & 0.20 & 0.20 & 1202.28 \\
& K-Center               & 15.44 & 0.34  & 0.37 & 91.86 & 0.16 & 0.16 & 1769.58 \\
& \cite{yoo2019learning} & 11.69 & 0.17  & 0.18 & 91.61 & 0.24 & 0.24 & \textcolor{white}{0}156.52 \\ \midrule
\multirow{4}{*}{\textsc{Redundancy-free}} 
& w/o selection          & 15.09 & -4.18 & 9.27 & 16.77 & 0.03 & 0.02 & 4471.01 \\ \cmidrule{2-9}
& Random                 & 12.28 & -2.43 & 6.36 & 16.70 & 0.03 & 0.03 & \textbf{5207.31} \\
& K-center               & 12.22 & -2.51 & 6.49 & 16.69 & 0.03 & 0.03 & 4912.63 \\
& \cite{yoo2019learning} & \textcolor{white}{0}9.82  & -2.80 & 6.34 & 16.72 & 0.03 & 0.04 & 1639.60\\ \bottomrule
\end{tabular}%
\end{threeparttable}
}
\caption{Comparison between different coreset selection methods with two stream settings. \textit{Random} represents the random selection for each mini-batch.}
\label{tab:cs}
\end{table*}


\begin{table}[t]
\centering
\small
\fontsize{7}{10}\selectfont

\label{tab:cs_r}
\resizebox{0.43\textwidth}{!}{%
\begin{tabular}{ccccccc}
\toprule
Ratio & EM & BWT & FWT & DTW & KAR \\ \midrule
r=0.25 & \textcolor{white}{0}9.60 & -1.94 & 4.74 & 16.68 & 4369.71 \\
r=0.50 & 12.22 & -2.51 & 6.49 & 16.69 & \textbf{4912.63} \\
r=0.75 & 14.13 & -3.99 & 8.23 & 16.64 & 4604.69 \\
r=1.00 & 15.09 & -4.18 & 9.27 & 16.77 & 4471.01 \\ \bottomrule
\end{tabular}%
}
\caption{Varying different coreset selection ratios based on the best coreset selection method.}
\label{tab:cs_r}
\end{table}

\begin{table}
\small
\centering
\fontsize{8.5}{12}\selectfont
\begin{threeparttable}
\resizebox{0.4\textwidth}{!}{
\begin{tabular}{lcccc}
\toprule
\multicolumn{1}{c}{Method} & EM & BWT & FWT & KG  \\ \midrule 
T5-Vanillla & \textcolor{white}{0}7.48 & \textbf{-0.84} & 4.07 & 40.53 \\
T5-Recadam & 10.05 & -2.86 & 6.28 & 16.73 \\
T5-Mixreview & 11.00 & -2.22 & \textbf{6.83} & 16.70 \\
T5-LoRa & \textbf{11.06} & -1.87 & 6.31 & 16.69 \\
T5-Kadapter & \textcolor{white}{0}9.88 & -1.96 & 6.60 & \textbf{16.57} \\
T5-KD & \textcolor{white}{0}2.13 & -3.20 & -0.94 & 40.57 \\
T5-Modular & 10.31 & -1.85 & 5.77 & \textbf{16.57} \\
\bottomrule
\end{tabular}%
}
\end{threeparttable}
\caption{Comparison of knowledge forgetting and updating between different methods in the time limitation setup.}
\label{tab:time limitation}
\end{table}

To attain an optimal balance between training velocity and model performance, we investigate the influence of coreset selection on OCKL in both redundant and non-redundant data stream scenarios. The performances of distinct coreset selection strategies are elaborated in Table~\ref{tab:cs}.

\paragraph{Redundancy} 
In the redundant data streams, the K-Center method emerges as the most effective, whereas random sampling is demonstrably inferior. Both K-Center and the predictive network approach proposed by \citet{yoo2019learning} leverage the repetitive nature of the redundant data stream. However, K-Center excels in isolating representative samples from the stream. It is noteworthy that none of the coreset methods manage to enhance the knowledge acquisition rate in the online setting.

\paragraph{Redundancy-free}
Contrastingly, in the absence of data redundancy, simple random sampling surpasses other methods in terms of both training speed and final performance. This can be ascribed to a misalignment between the methodological underpinnings of K-Center and the predictive network approach and the intrinsic properties of our data stream, particularly the continuous influx of distinct and evolving data. In both redundant and non-redundant settings, the Exact Match score experiences a decline. However, the knowledge acquisition rate sees an improvement in the non-redundant case, while deteriorating under redundant conditions. This can be attributed to a greater rate of information assimilation relative to forgetting in a redundancy-free environment, even though the forgetting rate remains substantial. Furthermore, as delineated in Table ~\ref{tab:cs_r}, varying coreset selection ratios in the K-Center method are explored, revealing improved performance as the ratio increases.


\subsubsection{Fast Knowledge Stream Arrival Rate}
In this work, drawing upon \citet{ghunaim2023real}, we established a scenario where the rate of incoming data stream surpasses the processing capacity of the current GPU cluster. This challenging yet achievable setting allowed us to assess the effectiveness of various continual learning strategies under time constraints. Unlike previous experiments focused on the continual learning capabilities of different methods, this investigation emphasizes a more practical scenario with time limitations. We enforced a uniform computation time across all methods. Consequently, any data exceeding this time frame was discarded. 

The outcomes of this scenario are detailed in Table~\ref{tab:time limitation}, where the time constraint is set at half the duration of Kadapter's processing time. This effectively means that, using Kadapter's processing speed as a benchmark, the model must omit half of the training data. Notably, LoRA and Mixreview outperformed under these conditions. Contrary to previous findings, LoRA exceeded Mixreview in performance, likely due to its inherently greater computational efficiency. Mix-Review exhibited outstanding FWT abilities and notable success in EM performance. Thus, we suggest that in real-world settings, the choice of a CL method should be contingent on the specific environmental demands. In cases where the rapid pace of data flow necessitates discarding data, methods like LoRA (parameter expansion), are recommended. Alternatively, when the system can process most of the data, Mix-Review tends to provide superior outcomes.

\begin{figure}[t]
\centering
\includegraphics[width=0.4\textwidth]{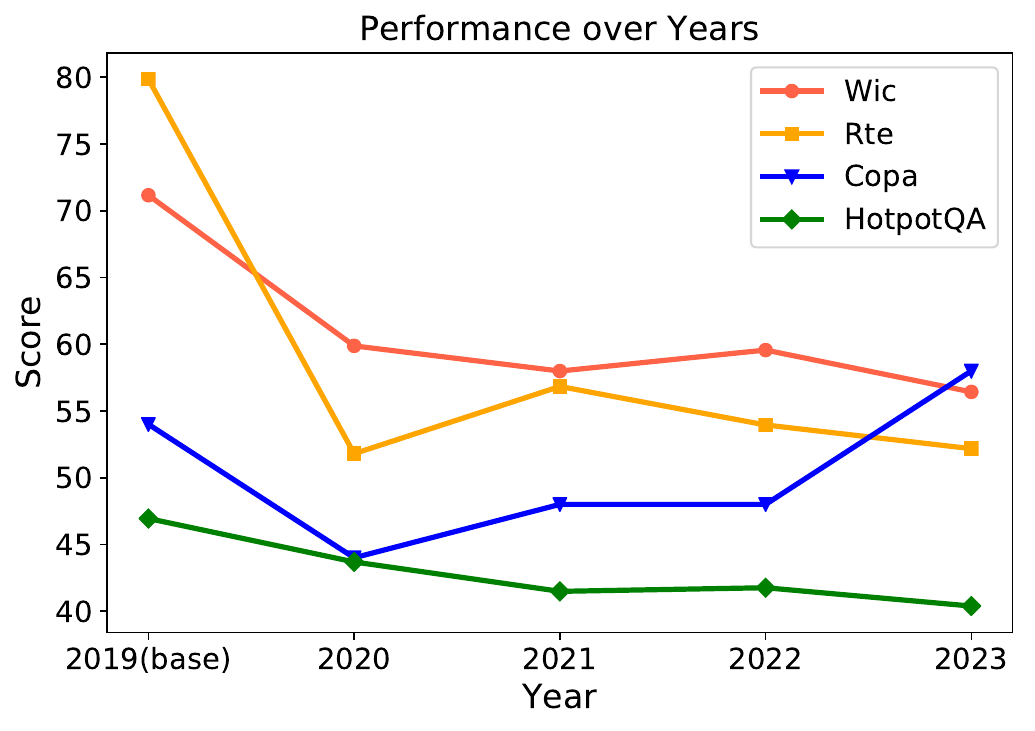}
    \caption{Changes in LM’s fundamental capabilities during the OCKL process.
}
    \label{fig:over time}
\end{figure}

\begin{table}[t]
\centering
\small\addtolength{\tabcolsep}{-0pt}
\fontsize{6}{8}\selectfont
\resizebox{0.49\textwidth}{!}{%
\begin{tabular}{lccccccc}
\toprule
\multicolumn{1}{c}{\textbf{Method}} & \textbf{EM} & \textbf{BWT} & \textbf{FWT} & \textbf{KG}  & \textbf{KAR} \\ \midrule
Initial & 0.217 & -- & -- & 0.278  & -- \\ \midrule
Vanilla & 0.192 & 0.000 & -0.014 & 0.853  & -4.996 \\
RecAdam & 0.115 & 0.011 & -0.004 & 0.609  & \textcolor{white}{0}1.191 \\
Mix-Review & 0.182 & 0.003 & -0.007 & 0.802  & -1.519 \\
LoRA & 0.242 & -0.005 & -0.002 & 0.887  & -3.180 \\
Kadapter & 0.180 & -0.032 & 0.020 & 0.311  & -5.589 \\ \bottomrule
\end{tabular}%
}
\caption{\small Failure performance of decoder-only models GPT2 initially pretrained on Dec 2019 dump of Webtext.}
\label{tab:gpt2}
\end{table}

\subsubsection{Impact of OCKL on Language Model Fundamental Abilities }
In this section, we focus on how continuous knowledge acquisition impacts the fundamental capabilities of LMs. To explore this, we have utilized LMs that has been subject to continual pre-training. Subsequently, we fine-tuned and evaluated this model to probe for any potential changes or improvements in its core abilities, including reasoning and discerning the meanings of words. Figure~\ref{fig:over time} presents the performance of the pre-trained models over different periods. It is observed that there is a general, albeit slight, decline in performance metrics, with the notable exception of the COPA dataset, which shows an initial dip followed by a partial recovery. In contrast, the decline in the HotpotQA dataset is relatively mild, indicating that the OCKL has a minimal impact on the model's reasoning capabilities. However, the RTE dataset exhibits a more significant drop, which may be attributed to the OCKL's more uniform input format, potentially impairing the LM's comprehension abilities. 

\subsubsection{Exploring How OCKL Methods Transfer Across LM Architectures}
\label{appen:gpt-2}
Following the approach of \cite{jang2021towards}, our experiments involved GPT-2 Large~\citep{radford2019language} of 774M parameters, which is initially pretrained on WebText (Dec 2019)\footnote{WebText comprises 8 million documents, Wikipedia excluded.}, and later trained on our dynamic knowledge stream for one epoch at a learning rate of 1e-4. The results, presented in Table \ref{tab:gpt2}, were not as anticipated, with all strategies underperforming compared to the baseline GPT-2. The underperformance in our GPT-2 Large experiments is mainly due to the rapid nature of online training and the model's decoder-only architecture. Online training restricts GPT-2's exposure to each batch of data, limiting its adaptability to the rapidly changing content typical of streaming data environments. Additionally, the model's focus on generating responses rather than deeply encoding inputs hampers its capacity to process complex, evolving information in our knowledge stream. The brevity of our textual data, rich in content but limited in length, further challenges the model. These insights point towards the need for modifications in both data handling and model architecture to enhance learning efficacy in dynamic online environments.

\section{Conclusion}
In this paper, we underline the imperative for real-time evaluation metrics in Continual Learning methods, with a particular focus on the continuous pre-training of Language Models. We unveil \textit{Online Continual Knowledge Learning}, a novel framework furnished with a dynamically adaptive benchmark dataset. We propose two innovative metrics: Knowledge Acquisition Rate and Knowledge Gap. Our empirical evaluation reveals that prevailing CL strategies are ill-suited for real-time, stream-based applications. With these insights, we urge the academic community to utilize \dataset as a foundational resource for the future refinement and innovation of adaptive, continually updating LMs.

\bibliography{anthology,custom}
\bibliographystyle{acl_natbib}
\newpage
\section{Appendix}
\subsection{Traditional Continual Learning}
\label{tranditional CL}
Continual learning methodologies have evolved to mitigate catastrophic forgetting, a core challenge in this domain. These strategies can be broadly categorized into three types: \textit{regularization}, \textit{replay-based}, and \textit{parameter expansion} methods. 

Regularization techniques, such as those proposed by \citet{buzzega2020dark} and \citet{caccia2021new}, aim to modify the classification objective to preserve previously learned representations, thus fostering more robust and versatile knowledge retention. Replay-based methods, detailed in works like \citet{chaudhry2019continual} and \citet{aljundi2019gradient}, prioritize the selection and retention of representative memory samples, applicable in both offline and online training settings. Parameter expansion strategies, explored by \citet{rusu2016progressive} and \citet{hu2018overcoming}, involve training specific task parameters while keeping others static, allowing for more focused learning.

Studies such as \citet{luu2021time} and \citet{dhingra2022time} have further enriched the field by examining the effects of temporal misalignment in learning. While \citet{luu2021time} delves into the impact of various tasks over extended periods, \citet{dhingra2022time} introduces a method for concurrently modeling text with its associated timestamp, enhancing the memorization of known facts and the calibration of predictions for unseen ones. Extending this approach, \citet{margatina2023dynamic} increased the granularity of evaluations to better understand LMs’ handling of evolving factual knowledge. However, these methods predominantly function in an offline capacity, requiring multiple passes over the data—a constraint that limits their practicality in rapidly changing data stream scenarios.


\subsection{Coreset Selection}
\label{coreset}
\paragraph{Random Selection}
A straightforward method for reducing the size of the training data is random selection, which involves randomly choosing a subset of samples from the original dataset. This method's principal advantage is its simplicity and computational efficiency, as it obviates the need for complex selection criteria or domain-specific heuristics. This makes it especially suitable for scenarios involving large datasets where computational resources and time are limited.

Nonetheless, it is essential to acknowledge the limitations of random selection. Specifically, this method may inadvertently exclude informative or representative samples that could be crucial for model training. Moreover, it is ill-suited for applications involving imbalanced datasets since it does not take the distribution of the data into account.

\paragraph{K-Center Selection}
The K-Center problem, a classical issue in combinatorial optimization and graph theory, has recently been applied innovatively in active learning~\citep{sener2017active,borsos2020coresets}. In this approach, K centroids are identified to form a coreset, thereby enhancing the quality of the selected data. Extending this methodology, \citet{li2022camel} introduced the $\delta-cover$ concept to efficiently prune redundant data in image data streams. This innovation sets an upper bound on the objective function without the need for gradient or loss value computations, effectively reducing training overhead due to redundant or less-informative data instances.

However, transitioning this approach to text data and leveraging embeddings for coreset selection introduces several challenges that we addressed in this work. One such issue is the potential for loss of nuanced details during the embedding process, which can compromise the accuracy of K-center selection. Such an omission can result from the inadequacy of the embedding space in capturing the subtleties inherent in the original data. Another challenge lies in the distortion of inter-point distances within the original dataset when translated into the embedding space, thereby affecting the efficacy of greedy K-center algorithms. This could lead to selections that do not accurately represent the intrinsic structure of the original data.

\paragraph{Model-Based Selection}
Employing model states, such as loss and gradients, serves as a viable strategy for the selection of representative samples for training. Importance sampling techniques~\citep{johnson2018training, katharopoulos2018not, sinha2022experience} have shown promise in hastening the training process of deep neural models. Specifically, \citet{yoo2019learning} introduced small trainable modules within the model architecture to predict the loss values for untrained samples. These modules enable the subsequent utilization of samples with low predicted loss values for further training. Complementing this, \citet{yoon2021online} employed gradient analysis to gauge sample similarity, diversity, and affinity, allowing for the informed selection of data samples in each training iteration.

While these approaches potentially reduce the dataset size and expedite training, it is imperative to consider the computational costs associated with such techniques. Tasks like the training of auxiliary modules and the computation of data-affinity coefficients add an additional computational layer, which must be weighed against the benefits of the accelerated training process.

\subsection{Data Construction Process}
\label{appen:datasets}
\begin{figure*}[!t]
\centering
    \includegraphics[width=0.7\textwidth]{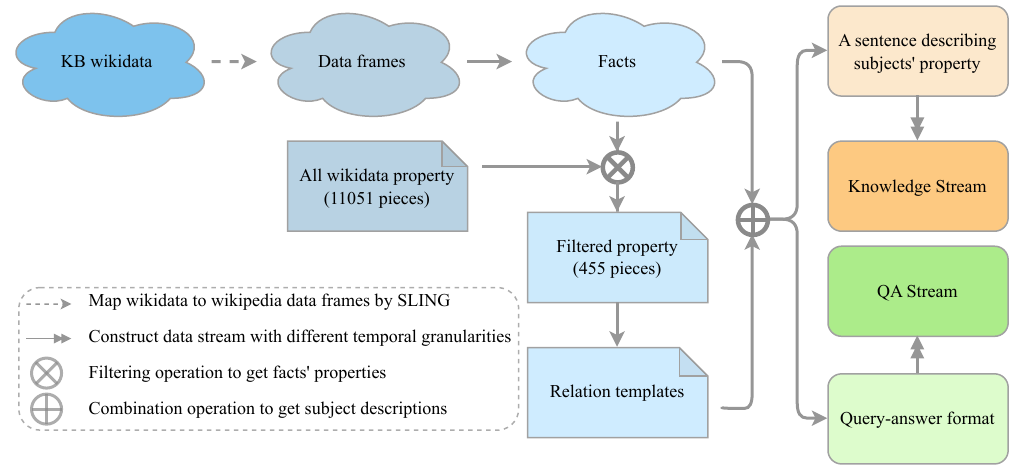}
    \caption{The process of constructing knowledge stream and QA stream for the training and evaluation.}
    \label{fig:data_pre}
\end{figure*}

A visual representation of our data construction process is shown in Figure~\ref{fig:data_pre}. Leveraging SLING, we utilize the extensive knowledge base of Wikidata to construct these streams. The construction process begins by mapping the Wikidata dump from 2019 to 2023 into multiple data frames. After being Filtered by the wikidata property and using the relation template, these structured frames are transformed into descriptive facts. These narratives are our \textit{Knowledge Stream} containing corpus and dates. For the \textit{QA Stream}, we using to a method that parallels previous works~\citep{margatina2023dynamic, dhingra2022time}. We convert factual narratives into question-answer formats thus structuring the \textit{QA Stream} with questions, answers, and dates for real-time evaluation. 

\subsection{Data Stream Analysis}
\label{append:analysis}
\begin{figure}[!t]
    \centering
    \subfigure[The CDF of the token change]{%
        \includegraphics[width=0.30\textwidth]{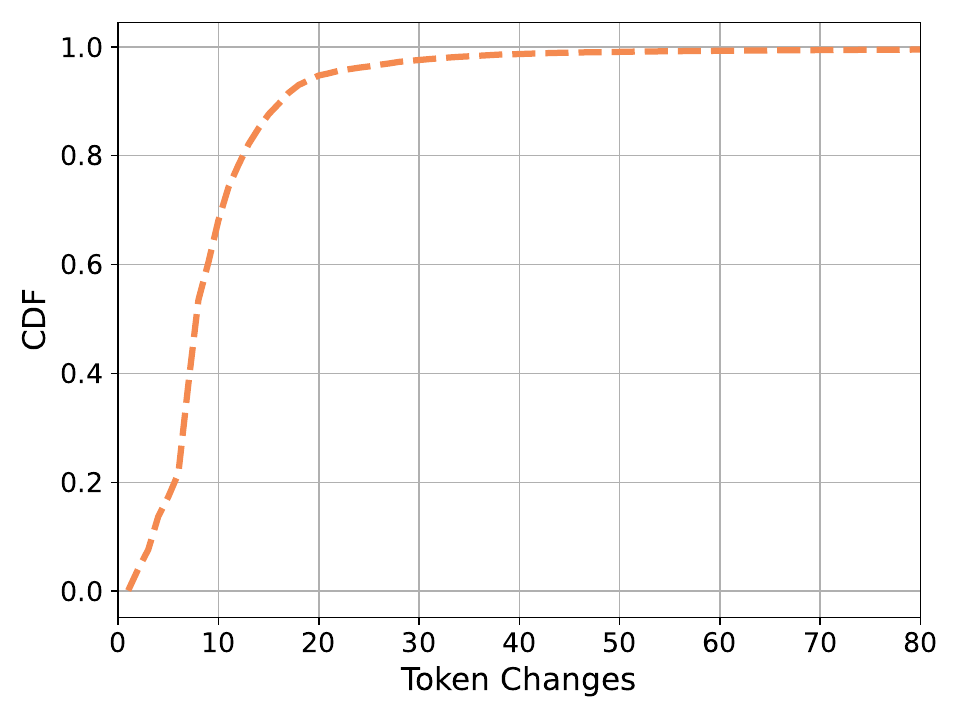}
        \label{fig:ana_tok}
    }
    \subfigure[The CDF of the date change]{%
        \includegraphics[width=0.30\textwidth]{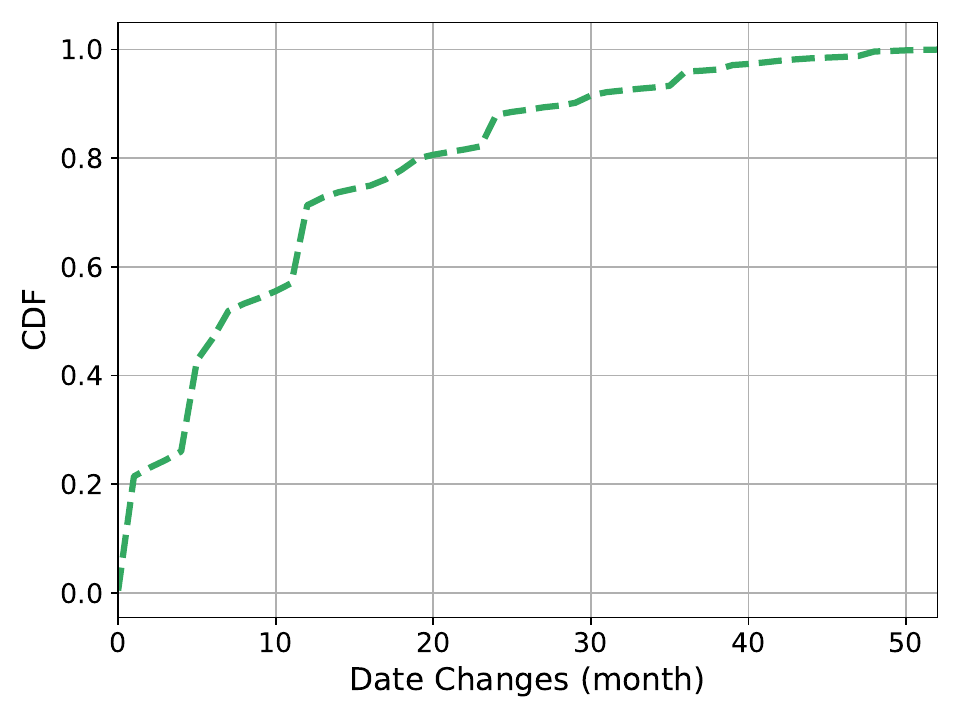}
        \label{fig:ana_date}
    }
    \caption{The Cumulative Distribution Function (CDF) of the token and date changes indicates the knowledge stream varies over time.}
    \label{fig:mainana}
\end{figure}

As shown in Figure \ref{fig:mainana}, 61.9\% of time-varying data has token changes of no less than 7, which indicates that most of the data have significant changes over time.

\subsection{CL Setups}
\label{CL setups}
\textbf{Vanilla} refers to a standard setup utilized for additional training, as discussed by~\citet{gururangan2020don}. In this context, the domain represents new knowledge. The Language Model (LM) undergoes further pretraining without the adoption of specialized training techniques, instead leveraging the original training strategy for incremental learning.

\textbf{RecAdam}~\citep{chen2020recall} falls into the category of regularization methods. It differs from traditional regularization methods like EWC~\citep{kirkpatrick2017overcoming} by imposing a more stringent independence assumption among model parameters. Notably, it doesn't use the initial pretraining corpus for regularization during continued training. The regularization strength diminishes as training advances, achieved by annealing the optimizer.

\textbf{Knowledge Distillation:} \citet{gou2021knowledge} introduced a widely-used method to speed up model inference by minimizing the representation gap between two models. Building on this concept, studies by \citet{8107520,rebuffi2017icarl,hou2018lifelong} have implemented distillation techniques in continuous learning. They do this by designating the model from an earlier task or domain as the teacher, and the model currently being trained as the student. This strategy effectively combats catastrophic forgetting. Recently, \citet{jin2021lifelong} expanded the use of knowledge distillation to the pre-training of language models, with encouraging outcomes. We extend this approach to the OCKL  with the subsequent abbreviation T5-KD.

\textbf{Mix-Review}~\citep{he2021analyzing} falls into the category of rehearsal methods, which assumes access to the initial pretraining corpus and mixes in random subsets of the initial pretraining data during continued pretraining, depending on the mix-ratio at the current time step. As the training process advances, the mix ratio gradually decreases toward 0, resulting in a reduced presence of the original mixed data with each iteration. 

\textbf{LoRA}~\citep{hu2021lora} is a parameter-expansion technique presented by~\citet{hu2021lora}. Instead of updating the original LM parameters, it introduces trainable rank-decomposition matrices in each layer for continued pretraining. While \citet{hu2021lora} applied this to decoder-only models like GPT-2~\citep{radford2019language}, \citet{jang2021towards} application extends to encoder-decoder models.

\textbf{K-Adapter}~\citep{wang2020k} is another parameter expansion method that freezes the original LM parameters while adding a \textit{  k} number of new layers, namely \textit{adapters}, that are updated during continued pre-training. \citet{wang2020k} have shown successful injection of \textit{factual} and \textit{linguistic} knowledge for encoder-only models, BERT~\citep{devlin2018bert} \& RoBERTa~\citep{liu2019roberta}, while we also apply it to an encoder-decoder model, T5, and decoder-only model, GPT-2.

\textbf{Modular}~\citep{jang2021towards} is a novel parameter expansion strategy tailored for encoder-decoder models. The original pretrained encoder is preserved, but a supplementary, randomly initialized encoder is introduced for updates during the next phase of training. 

\subsection{Hyperparameter and Device Settings}
\label{Hyper-parameter}
Table \ref{tab:hparam} is our hyperparameter settings and GPU device.

\begin{table}[t]
\centering
\small
\fontsize{4}{5}\selectfont
\resizebox{0.35\textwidth}{!}{%
\begin{threeparttable}
\begin{tabular}{lr}
\hline
\multicolumn{1}{c}{Hyper-parameters} & \multicolumn{1}{c}{Value}\\ \hline
Input length & 18 \\
Output length & 18 \\
epoch & 1\\
Training batch size & 8 \\
Testing batch size & 512 \\
Model & T5-base \\
Optimizer & Adafactor \\
Weight decay & 5e-04 \\
\# of GPUs & 1 \\
GPU & A40 48GB \\
Learning rate & 2e-03 \\ \hline
\end{tabular}%
\end{threeparttable}
}
\caption{Hyper-parameters for T5 experiments}
\label{tab:hparam}
\end{table}

\subsection{Evaluation Metrics}
\label{other Metrics}
\paragraph{Exact Match (EM)}
The ``Exact Match'' (EM) metric is employed to assess the fidelity of the model's generated answers in the Question-Answering (QA) context. An answer is designated as an ``exact match'' if it perfectly aligns with the ground-truth answer in terms of wording, sequence, and punctuation. The mathematical formulation for this metric is as follows:
\begin{equation}
EM = \frac{{\text{Number of exact matches}}}{{\text{Total number of examples}}}
\end{equation}

\paragraph{Backward Transfer (BWT)\label{BWT}}
This metric quantifies the performance degradation on prior tasks after a model has been trained on new tasks, a phenomenon known as "forgetting" in the context of continual learning. Traditional BWT approaches~\citep{lopez2017gradient} compute the mean performance decrement across all preceding tasks after the acquisition of new tasks, offering a generalized measure of the model's retention abilities. To delve deeper into the problem of forgetting, particularly with respect to tasks that are temporally adjacent, we introduce an alternate formulation for BWT as follows:
\begin{equation}
\text{BWT} = \frac{1}{T-1} \sum_{i=2}^{T} (R_{i,(i-1)} - R_{(i-1),(i-1)}),  \label{eq:bwt}
\end{equation}
where \( R_{i,(i-1)} \) denotes the test accuracy of the model trained on tasks \( [1,i] \) when evaluated on the preceding \( i-1 \) tasks. This adapted metric allows us to focus on performance shifts related to forgetting in consecutive tasks. Given that our data stream exhibits temporal continuity and task correlation—e.g., tasks may share hierarchical or categorical relationships—this revised metric furnishes a more nuanced understanding of how the model's performance evolves, especially concerning the forgetting issue, as it transitions from one task to another.

\paragraph{Forward Transfer (FWT)\label{FWT}}
This metric evaluates the positive impact on the performance of new tasks when a model has completed training on preceding tasks. Traditional FWT approaches~\citep{lopez2017gradient} assess the performance differential on a new task between models that have been trained on all prior tasks and a baseline model that has not undergone any training. While this method offers an aggregate view of how prior learning influences new tasks, it lacks granularity in task-specific contexts. To address this limitation, we introduce an alternative formula for FWT as follows:
\begin{equation}
\text{FWT} = \frac{1}{T-1} \sum_{i=2}^{T} (R_{i,i} - R_{i-1,i}),  \label{eq:fwt}
\end{equation}
where \( R_{i,i} \) represents the test accuracy of the model trained on tasks \( [1,i] \) when evaluated on the new \( i \)-th task. This refined metric enables us to focus on the performance increments related to task-specific updates in consecutive tasks. Given the temporal continuity and task correlations in our data stream, the new metric serves to better scrutinize how the acquisition of new knowledge influences or disrupts model learning at each transitional phase. This approach thus provides a more nuanced, fine-grained evaluation of the model's ability to integrate new knowledge without adverse effects. To further enhance the depth of our analysis, we also employ vector representations to quantify the updating issue, offering a more comprehensive evaluation of knowledge assimilation.



\end{document}